\documentclass{article}

\usepackage{PRIMEarxiv}

\usepackage[utf8]{inputenc} 
\usepackage[T1]{fontenc}    
\usepackage{hyperref}       
\usepackage{url}            
\usepackage{booktabs}       
\usepackage{amsfonts}       
\usepackage{nicefrac}       
\usepackage{microtype}      
\usepackage{lipsum}
\usepackage{fancyhdr}       
\usepackage{graphicx}       
\graphicspath{{media/}}     
\usepackage{wrapfig}
\usepackage{amsmath}

\title{ Vertical LoRA: Dense Expectation-Maximization Interpretation of Transformers
}

\author{
  Zhuo-Lin Fu \\
  \texttt{zfu19@hawk.iit.edu} \\
}

\begin{document}
\maketitle

\begin{abstract}
In this paper, we show how Transformers can be interpreted as dense Expectation-Maximization algorithms performed on Bayesian Nets.
Based on the above interpretation, we propose a new model design paradigm, namely Vertical LoRA (VLoRA), which reduces the parameter count dramatically while preserving performance.
In VLoRA, a model consists of layers, each of which recursively learns an increment based on the previous layer.
We then apply LoRA decomposition to the increments.
VLoRA works on the base model, which is orthogonal to LoRA, meaning they can be used together.
We do experiments on various tasks and models.
The results show that 1) with VLoRA, the Transformer model parameter count can be reduced dramatically and 2) the performance of the original model is preserved.
The source code is available at \url{https://github.com/neverUseThisName/vlora}
\end{abstract}


\section{Introduction}
In recent years, the field of machine learning, especially natural language processing (NLP), has witnessed a transformative evolution, primarily catalyzed by the advent of Transformer models and large language models.
These models are known for their emergent ability to comprehend and generate human-like text.
Specifically, Transformer models seem to undergo a transformative evolution with the growth of parameter count, achieving unprecedented performance across a spectrum of tasks, including text generation, machine translation, text summarization, question answering, and visual understanding.
This finding leads to the trends in scaling up models up to millions and even Billions of parameters, exemplified by OpenAI's GPT\cite{radford2019language, brown2020language}, Google's BERT\cite{devlin2018bert}, Meta's Llama\cite{touvron2023llama}, and Anthropic's Claude\cite{Anthropic2024The}.

However, this scaling in model size simultaneously has rendered a significant barrier for ordinary individuals to train these models on consumer hardware setups.
For example, training Meta's Llama 3 model took a cluster with 24000 GPUs, which is prohibitively expensive for small institutes and individual users.

There have been mainly 3 lines of research to alleviate this difficulty: parameter compression\cite{dettmers2024qlora}, efficient model design\cite{hu2021lora, houlsby2019parameter, pfeiffer2020adapterfusion, zhang2023adaptive, wang2020linformer}, and efficient algorithms\cite{li2021prefix, liu2021p, lester2021power, kitaev2020reformer}.
This work falls into the category of efficient model design.
Amongst the line of efficient model design, there is a branch of research using low-rank decomposition to reduce parameter count while preserving performance.

LoRA\cite{hu2021lora} is of particular relevancy to this work.
Given a pre-trained Transformer model, LoRA fine-tunes the model by learning an increment to each layer's pre-trained weights and then using two low-rank weight matrices to factorize and approximate the increment.
Note LoRA works in the fine-tuning phase and each layer learns an increment independently.

In the paper, we first claim that Transformer models trained with supervision are effectively Expectation-Maximization (EM) algorithms that maximize the posterior $P(y|x;\theta)$.
Each layer of a Transformer model is effectively an iteration of such an EM algorithm, where the forward pass corresponds to the E step, and the weight difference of the next layer from the current layer implicitly corresponds to the M step.
Based on the above claim, we further claim that each layer of a model effectively learns an increment based on the previous layer.
Finally, we propose a new model design paradigm, named Vertical LoRA (VLoRA): we first define a full-rank base layer, and then the second layer only learns an increment based on the base layer; this process goes recursively, i.e. the third layer increments from the second and so on.

It is worth noting that there's a recent work LORS\cite{li2024lors} that's very similar to this work.
We discuss its difference from this paper in Section \ref{subsec:LORS}.

\section{Related Work}
\label{sec:related_work}

\subsection{Low-rank Decomposition}
The weight matrices of deep learning models were found to be redundant in the number of parameters\cite{denil2013predicting}.
Subsequently, there has been a line of research devoted to finding the effective rank of weight tensors\cite{aghajanyan2020intrinsic, li2018measuring} and reducing the number of parameters by approximating the original tensor by low-rank ones\cite{jaderberg2014speeding, sainath2013low, zhang2015accelerating, xue2013restructuring, denton2014exploiting, lebedev2014speeding, kim2015compression}.

\subsection{LoRA}
Low-Rank Adaptation of Large Language Models (LoRA), inspired by the above work, is an approach proposed to address the computational and memory challenges associated with fine-tuning large language models (LLMs) for downstream tasks.
Traditional full fine-tuning of LLMs requires substantial computational resources and memory due to the vast number of parameters involved.
LoRA introduces a low-rank factorization technique to reduce the computational and memory overhead while preserving model performance.

Specifically, let $\theta^{(l)}_{\text{pre-train}}$ be the pre-trained attention weight matrix of layer $l$. LoRA fixes $\theta^{(l)}_{\text{pre-train}}$ and only trains an increment $\Delta\theta^{(l)}$, and adds it to the pre-trained weights
\begin{equation}
    \theta^{(l)}_{\text{finetune}} = \theta^{(l)}_{\text{pre-train}} + \Delta\theta^{(l)}
\end{equation}
and then factorizes the increment into two low-rank components $A$ and $B$
\begin{equation}
    \Delta\theta^{(l)} = B^TA.
\end{equation}
LoRA effectively reduces the number of parameters and the computational complexity of the fine-tuning process, resulting in faster training and lower memory requirements.

There has been a rich proliferation of LoRA family.
To name a few, AdaLoRA\cite{zhang2023adaptive} decomposes the incremental matrix with SVD and enforces orthogonality with a regularizer term.
It further dynamically masks out small eigenvalues to adaptively allocate rank budget on each matrix.
PeriodicLoRA\cite{meng2024periodiclora} proposes a multi-stage LoRA, where a new pair of low-rank $A$ and $B$ are learned in each stage and offloaded to the pre-trained weights.
LoRA Dropout\cite{lin2024lora} applies dropout regularization to the rows of $B$ and columns of $A$ and achieves better test performance.
LoRA+\cite{hayou2024lora+} claims that the default LoRA training with equal learning rates on $A$ and $B$ is inefficient.
They propose a scheme to assign different learning rates to them, along with a parameter initialization method.

LoRA family works in the post-training (fine-tuning) phase, and it reduces trainable parameters only for fine-tuning.
LoRA factorizes each layer independently from other layers, i.e. in the "horizontal" direction.
This paper (VLoRA) works in the realm of efficient model design and reduces the overall parameters for the base models.
VLoRA factorizes each layer recursively, where each layer is factorized based on the previous layer, i.e. in the "vertical" direction.

\subsection{LORS}
\label{subsec:LORS}
LORS, which stands for Low-rank Residual Structure for Parameter-Efficient Network Stacking, is an approach proposed to reduce the overall model parameters.
LORS defines a weight matrix $\theta_{\text{shared}}$ shared by all layers and a private weight matrix $\theta^{(l)}_{\text{private}}$ for every layer $l$.
The total weights of layer $l$ is 
\begin{equation}
    \theta^{(l)} = \theta_{\text{shared}} + \theta^{(l)}_{\text{private}}
\end{equation}
Then the private matrix is factorized by a group of $2K$ low-rank matrices.
\begin{equation}
    \theta^{(l)}_{\text{private}} = \sum_{i=1}^K B_{l,i}A_{l,i}.
\end{equation}
Our work differs from it in the following ways: 1) this work provides theory support (the EM algorithm) for such motivation; 2) this work factorizes layers recursively (a layer depends on the previous layer), while layers in LORS are independent of each other; 3) our theory also leads to a chunking model design to account for the possible hierarchical structure in the latent space.

\section{Method}
\label{sec:method}

\subsection{Transformers as EM Algorithms}
In this subsection, we explain why and how we can pose Transformer models as dense Expectation-Maximization algorithms.
For simplicity, we consider an encoder Transformer model trained for a discriminative task (e.g. classification).
We first illustrate the idea with examples in both vision and language.

\textbf{A vision example}.
In a Vision Transformer, an input image is partitioned into disjoint patches.
The task of the Vision Transformer is to map the input to its label, which often entails inferring the latent variables (a.k.a features) for each patch.
However, making inferences on the latent variables of one patch often relies on those of the neighboring patches.
For example, in order to decide that patch 1 is from a nose, knowing that its neighboring patches are respectively from the eyes, mouth, and cheeks is very helpful.
This holds true for any other patches.

\textbf{A language example}.
In the phrase "New York", in order to determine that "New" is generated by a place, knowing "York" is also from a place is crucial, and vice versa.

This interlocking mechanism makes inference difficult.
This is also where the EM algorithm comes into play.

In fact, we can pose Transformers (and other DNNs) as the EM algorithms aimed at making such inferences.
Specifically, we can view a layer of a Transformer as an iteration of the EM algorithm.
The forward pass of a layer corresponds to the E step of the EM algorithm, where the latent variables for each patch are updated and refined based on those from last layer.
The change in weights of the next layer compared to the current layer implicitly corresponds to the M step, where the model weights are updated.

Formally, consider supervised learning.
Let $x$ be the input, $y$ the label, $z$ the unobserved latent variables, and $\theta$ the model parameters.
The goal of the model is
\begin{equation}
    \arg\max_\theta P(y|x;\theta)
\end{equation}
We have 
\begin{align}
&\log P(y|x;\theta) = \log P(y,z|x;\theta) - \log P(z|y,x;\theta)
\\
\Rightarrow
& \mathbb{E}_{z\sim P(z|y,x;\theta^{(l)})}\log P(y|x;\theta)
= \mathbb{E}_{z\sim P(z|y,x;\theta^{(l)})}\left[\log P(y,z|x;\theta) - \log P(z|y,x;\theta)\right]
\\
\Rightarrow
&\log P(y|x;\theta)
= \mathbb{E}_{z\sim P(z|y,x;\theta^{(l)})}\log P(y,z|x;\theta) - \mathbb{E}_{z\sim P(z|y,x;\theta^{(l)})}\log P(z|y,x;\theta),
\end{align}
where $\theta^{(l)}$ is the parameter from last iteration.

Let $Q(\theta|\theta^{(l)})=\mathbb{E}_{z\sim P(z|y,x;\theta^{(l)})}\log P(y,z|x;\theta)$ and $H(\theta|\theta^{(l)})=-\mathbb{E}_{z\sim P(z|y,x;\theta^{(l)})}\log P(z|y,x;\theta)$. We have
\begin{equation}
    \log P(y|x;\theta)
    = Q(\theta|\theta^{(l)}) + H(\theta|\theta^{(l)}).
\end{equation}
Then, 
\begin{align}
    &\log P(y|x;\theta) - \log P(y|x;\theta^{(l)}) \\
    = &Q(\theta|\theta^{(l)}) - Q(\theta^{(l)}|\theta^{(l)}) + H(\theta|\theta^{(l)}) - H(\theta^{(l)}|\theta^{(l)}) \\
    \ge &Q(\theta|\theta^{(l)}) - Q(\theta^{(l)}|\theta^{(l)}),
\end{align}
which indicates improvement in $Q(\theta|\theta^{(l)})$ results in at least as much improvement in log posterior. 

\begin{wrapfigure}{R}{0.25\textwidth}
    \centering
    \includegraphics[width=0.22\textwidth]{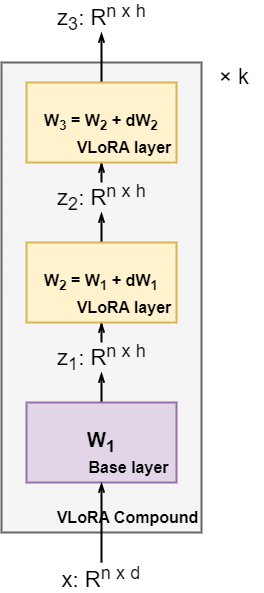}
    \caption{VLoRA model architecture. We partition a $L$-layer Transformer into $k$ chunks.
Each chunk contains $L/k$ layers; the first layer in a chunk is a base layer, and the remaining are VLoRA layers.}
    \label{fig:vlora_model_arch}
\end{wrapfigure}

The forward pass of a Transformer layer with weights $\theta^{(l)}$ corresponds to computing $Q(\theta|\theta^{(l)})$, i.e. the E step. The weights $\theta^{(l+1)}$ of the next layer correspond to the parameter $\theta$ that maximizes $Q(\theta|\theta^{(l)})$, i.e. the M step.
In Transformer, since each patch has its separate latent variable during inference (unlike CNN with strided convolutions), it is a dense EM algorithm. Training a model is equivalent to learning an EM algorithm.

The EM interpretation also explains the effectiveness of ResNet\cite{he2016deep}, where each layer is only responsible for learning an increment to latent variables.

\subsection{VLoRA}

The above observations lead to a natural paradigm for model design, named Vertical LoRA.
Since $\theta^{(l+1)}$ is updated based on $\theta^{(l)}$, we can factor it as
\begin{equation}
    \theta^{(l+1)} = \theta^{(l)} + \Delta\theta^{(l)},
\end{equation}
or 
\begin{equation}
    \theta^{(l+1)} = \theta^{(0)} + \sum_{j=0}^l\Delta\theta^{(j)}.
\end{equation}

This is where LoRA comes into play.
First, we define a full rank base parameter $\theta^{(0)}$.
Then similar to LoRA, we assume that each increment $\Delta\theta^{(l)}$ is of low rank, and we can apply LoRA decomposition to them:
\begin{equation}
    \Delta\theta^{(l)} = B^TA,
\end{equation}
where $A$ and $B$ are low rank matrices.
We call the layer with the base parameter the base layer and the layer with LoRA decomposition the VLoRA layer.

We refer to the original LoRA as horizontal LoRA, because it adapts every layer of a pre-trained model independently.
In contrast, VLoRA adapts each layer recursively in the "vertical" direction.

To account for the possible hierarchical structure in the latent variables, we partition a model with $L$ layers into $k$ chunks.
Each chunk contains $L/k$ layers; the first layer in a chunk is a base layer, and the remaining are VLoRA layers.
We call such a chunk a VLoRA Compound and the act of converting a Transformer model to its VLoRA version VLoRAfying the model.
The final model design is shown in Figure \ref{fig:vlora_model_arch}.

\section{Experiments}
In this section, we describe the experiments done and show the results.
All experiments are done with Transformer models.
The objective of the experiments is to compare the performance of Transformer models and their VLoRA versions by measuring performance on specific tasks.

A typical Transformer layer mainly consists of 4 sets of weights: QKV matrix, output projection matrix, and 2 projection matrices in the feed-forward module.
We VLoRAfy all of the 4 sets of weights.
We also experiment with different low-rank $r\in\{2,4,8\}$.

\subsection{Image Classification}
We do experiments on image classification tasks and use CIFAR-10\cite{krizhevsky2009learning} dataset.
We train from scratch a 12-layer Vision Transformer\cite{dosovitskiy2020image} and its VLoRA versions with different ranks.
The CIFAR-10 image size is $32\times 32$.
So we set the input patch size to $2\times 2$, the hidden dimension to 256, and the number of heads to 4.
For VLoRA, we empirically choose a moderate number of chunks $k=3$.
The comparison of the number of trainable parameters is shown in Table \ref{tab:vit_params}.

\begin{figure}[h]
    \centering
    \includegraphics[width=0.7\textwidth]{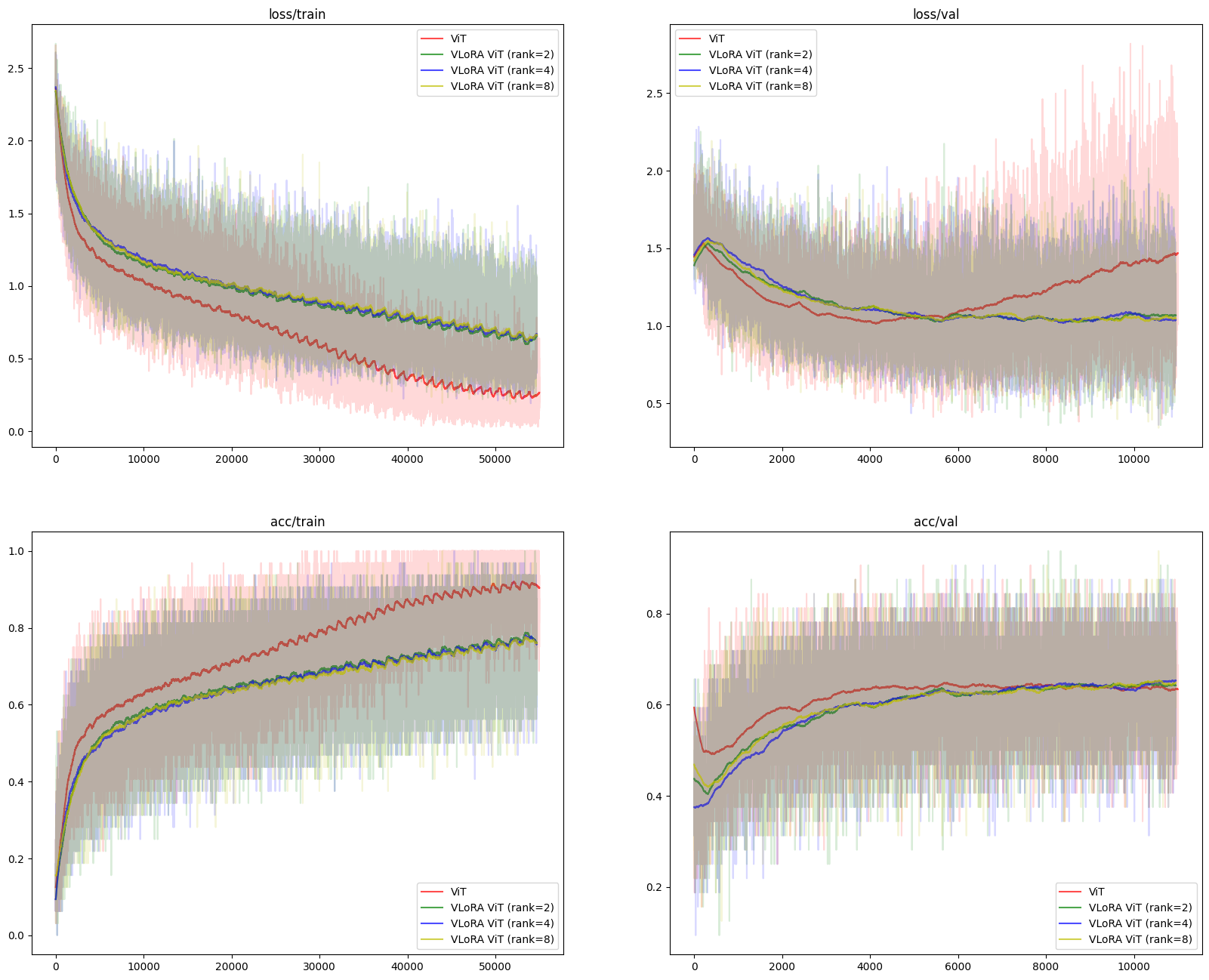}
    \caption{Training and evaluation loss and accuracy curves of ViT and its VLoRA versions on CIFAR-10}
    \label{fig:cifar10_train_graphs}
\end{figure}

\begin{table}
 \caption{Relative number of trainable parameters of VLoRA ViTs compared to the vanilla ViT. The ViT model hyperparameters: number of layers = 12, patch size = $2\times 2$, number of heads = 4, and hidden dimension = 256. The number of chunks for VLoRA is $k=3$}
  \centering
  \begin{tabular}{ccccc}
    \toprule
         &  ViT  & VLoRA ViT (rank=8) & VLoRA ViT (rank=4) & VLoRA ViT (rank=2) \\
    \midrule
    \#parameters (\%) & 100\%  & 28.97\% & 27.43\% & 26.65\%   \\
    \bottomrule
  \end{tabular}
  \label{tab:vit_params}
\end{table}
The training/evaluation loss and accuracy curves during training are shown in Fig.\ref{fig:cifar10_train_graphs}.

From the figure, We can see:
\begin{itemize}
    \item The training loss and accuracy of ViT is much better than its VLoRA versions.
    \item The evaluation loss and accuracy of ViT keep improving until some point in training and then take a turn for the worse.
    \item The evaluation loss and accuracy of VLoRA ViTs keep improving until they plateau.
    There is no significant turning point after which the metrics start to get worse.
    \item The evaluation performance of ViT is better at the beginning of training and is then overtaken by VLoRA ViTs.
    However, their best evaluation metrics are almost the same.
    \item The overall performances of VLoRA ViTs of different ranks are almost the same.
\end{itemize}
From the above observations, we can conclude:
\begin{itemize}
    \item VLoRA makes models less prone to overfitting.
    \item VLoRA preserves the performance of the original models.
    \item A small low rank (e.g. $r=2$) is sufficient to model the weight increment of each layer.
\end{itemize}

\section{Conclusion}
In this paper, with examples and solid theories, we show how we can pose Transformers as EM algorithms.
Based on this, we further propose a new model design paradigm that dramatically reduces parameter count while preserving the performance of the original models.
Experiments show that VLoRA not only reduces the parameter count but also performs better than the original models.

\bibliographystyle{unsrt}  
\bibliography{references}

\end{document}